\documentclass[10pt,twocolumn,letterpaper]{article}

 \usepackage{cvpr}              
\usepackage[table]{xcolor}
\definecolor{cvprblue}{rgb}{0.21,0.49,0.74}
\usepackage[pagebackref,breaklinks,colorlinks,allcolors=cvprblue]{hyperref}

\def\paperID{} 
\def\confName{CVPR}
\def\confYear{2026}

\title{SwitchCraft: Training-Free Multi-Event Video Generation with Attention Controls}

\author{
Qianxun Xu$^{1,2}$ \quad 
Chenxi Song$^{1, \dagger}$ \quad 
Yujun Cai$^{3}$ \quad 
Chi Zhang$^{1, *}$ \\
\vspace{0.1cm} \\
$^{1}$Westlake University \quad 
$^{2}$Duke Kunshan University \quad 
$^{3}$The University of Queensland \\
}

\usepackage{graphicx}
\usepackage{caption}

\begin{document}

\twocolumn[{%
  \maketitle
  \begin{center}
    \captionsetup{type=figure}
    \includegraphics[width=\linewidth,trim=6pt 6pt 6pt 6pt,clip]{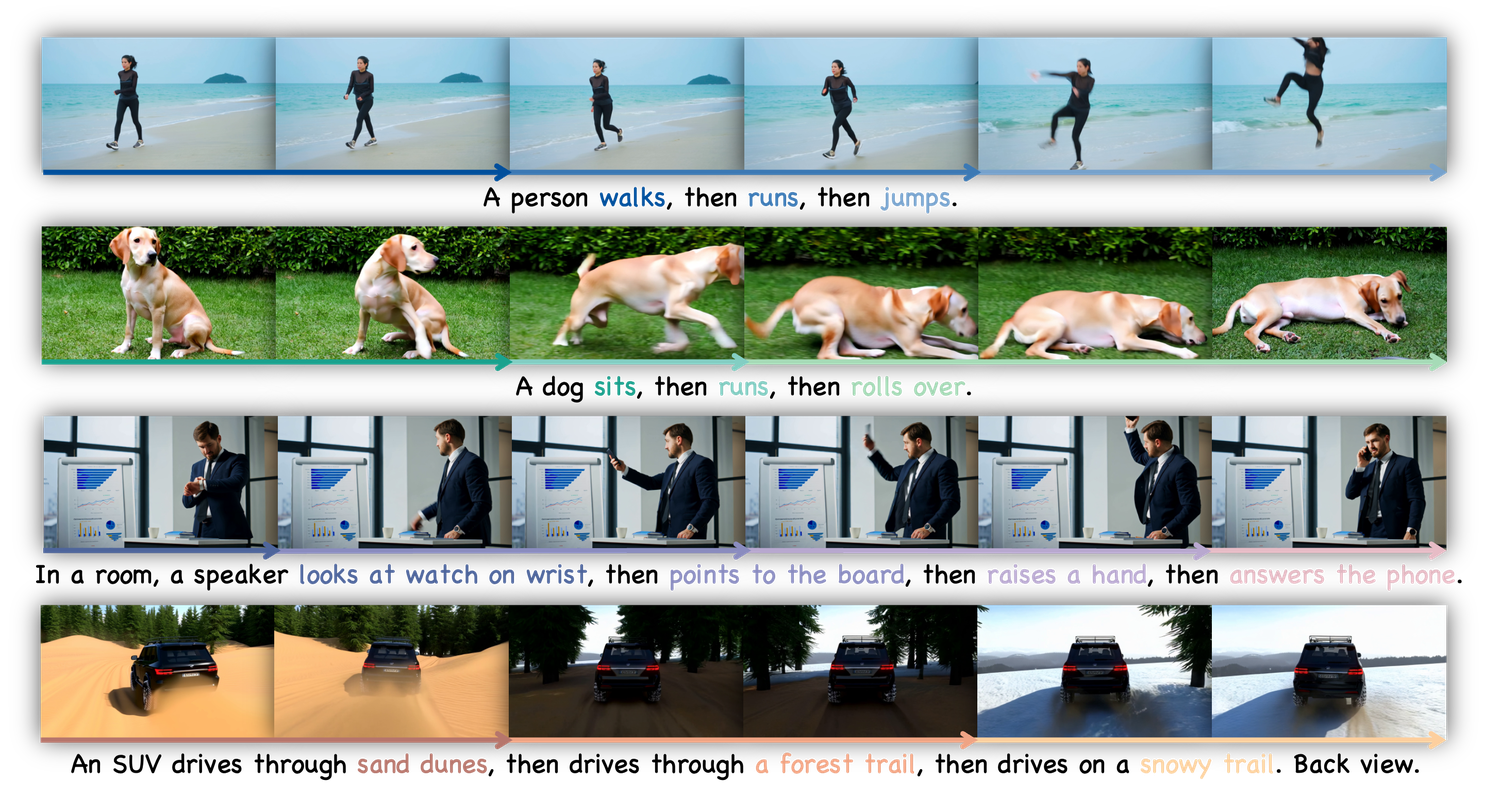}
    \captionof{figure}{\textbf{SwitchCraft} enables flexible multi-event video generation across multiple actions and scenes with smooth transitions. It steers attention to enhance prompt alignment and maintain coherent temporal evolution while preserving global information.}
    \label{fig:teaser}
  \end{center}
}]
\def\thefootnote{}\footnotetext{$^\dagger$Project lead. $^*$Corresponding author.}
\def\thefootnote{\arabic{footnote}}

\vspace{-0.8 cm}
\begin{abstract}

Recent advances in text-to-video diffusion models have enabled high-fidelity and temporally coherent video synthesis. However, current models are predominantly optimized for single-event generation. When handling multi-event prompts, without explicit temporal grounding, such models often produce blended or collapsed scenes that break the intended narrative. To address this limitation, we present SwitchCraft, a training-free framework for multi-event video generation. Our key insight is that uniform prompt injection across time ignores the correspondence between events and frames. To this end, we introduce Event-Aligned Query Steering (EAQS), which steers frame-level attention to align with relevant event prompts. Furthermore, we propose Auto-Balance Strength Solver (ABSS), which adaptively balances steering strength to preserve temporal consistency and visual fidelity. Extensive experiments demonstrate that SwitchCraft substantially improves prompt alignment, event clarity, and scene consistency compared with existing baselines, offering a simple yet effective solution for multi-event video generation. Code will be released \href{https://switchcraft-project.github.io}{here}.

\end{abstract}
\vspace*{-0.5 cm} 
\section{Introduction}
\label{sec:intro}

Recent years have witnessed remarkable progress in text-to-video (T2V) generation powered by diffusion models. Large-scale pretrained systems can now synthesize visually compelling video clips that align well with textual descriptions, exhibiting strong visual fidelity, temporal coherence, and identity consistency across diverse scenes~\cite{Ho2022VDM,Ho2022ImagenVideo,HunyuanVideo2024,wan2025}. These advances have significantly democratized video generation, making it central to a broad spectrum of applications~\cite{Wu2023TuneAVideo,Khachatryan2023Text2VideoZero,Lee2024GridT2V,Wang2024TFT2V,Jeong2024VMC,Jiang2024VideoBooth,Girdhar2024EmuVideo,BarTal2024Lumiere,Oh2024MEVG,Wang2025LaVie}.

Despite these advances, current state-of-the-art T2V models are predominantly optimized for generating single-event videos guided by one global caption prompt. In other words, they excel when the prompt describes a single coherent scene or action. 
However, when the prompt specifies multiple temporally ordered events,  
existing models often struggle to capture the compositional and temporal structure implied by the text. Instead of generating a sequence that faithfully follows the described narrative, the resulting videos tend to exhibit events blending, ambiguous transitions, or omit events and collapse into dominant scenes.
This limitation stems from how current architectures process text guidance during video generation. Typically, they apply the same caption representation uniformly across all timesteps through mechanisms such as cross-attention. As a result, the model interprets the entire description as a holistic context rather than a temporally structured sequence of events. This uniform conditioning neglects the intrinsic temporal hierarchy in complex prompts, making it difficult for the model to reason about event boundaries, causal order, or evolving scene dynamics.

Existing studies have attempted to introduce temporal control from two main directions. The first relies on training or fine-tuning the generation backbone with temporally grounded supervision, where each event is explicitly aligned with a specific time span~\cite{Wu2025MinT,Yang2025LongLive}. While this improves temporal correspondence between actions and frames, it requires densely annotated data, incurs heavy computational costs, and generalizes poorly to unseen or loosely structured scenarios. The second line adopts a concatenation strategy that fuses separately generated clips, leveraging information from previous segments to create smoother transitions~\cite{Cai2025DiTCtrl,Oh2024MEVG,Wang2023GenLVideo,Villegas2022Phenaki}. However, this clip-wise process lacks global context, so each segment has no foresight of subsequent events and cannot stage objects, poses, or motion cues in advance. Transitions thus become discontinuous and exhibit temporal drift across segments. 
Consequently, these approaches still fall short of achieving flexible multi-event video generation, where multiple actions evolve coherently within a single continuous generation process, maintaining temporal order and scene consistency without retraining the model.

Motivated by these limitations, we aim to address the challenge of multi-event video generation by optimizing how the model attends to different events over time. We identify that the core issue lies in the way prompts are injected through cross-attention: existing models lack a mechanism to accurately associate each event prompt with its corresponding temporal segment in the video. Without such alignment, attention becomes globally entangled, causing interference between unrelated events.

To overcome this, we introduce Event-Aligned Query Steering (EAQS), a training-free mechanism that aligns frame level attention with the corresponding event description. Instead of uniformly injecting the entire text prompt across all frames, EAQS modulates the query vectors of each frame using compact event-specific key subspaces built from textual anchor tokens. It increases the projection of queries onto the target event subspace and suppresses components aligned with competing events, which reduces event interference and improves disentanglement without changing model weights or training.

The steering strength is crucial for temporal control. Overly aggressive query perturbations may distort subject appearance or destabilize motion, whereas insufficient steering can fail to override the inherent bias of pretrained T2V models toward globally blended scenes. To robustly balance these competing factors, we propose the Auto-Balance Strength Solver (ABSS), an inference-time optimization module that adaptively computes enhancement and suppression coefficients for EAQS. ABSS compresses event-level directions via singular value decomposition, analyzes the current query–key alignment margins between the target event and all competing events, and solves a small convex optimization problem to obtain scale aware strengths that enlarge the target margin while limiting query displacement. This yields stable steering that preserves temporal coherence and visual fidelity with no additional hyperparameters or prompt-specific tuning.

To validate the effectiveness of our approach, we conduct extensive experiments on multiple benchmarks. The results demonstrate that our method significantly outperforms both baseline and state-of-the-art models in generating coherent, temporally ordered multi-event videos while maintaining high visual quality.
Our contributions are summarized as follows:
\begin{itemize}
    \item We propose \textbf{SwitchCraft}, a training-free framework enabling time-controlled multi-event video generation with clear event ordering and smooth transitions.
    
    \item  We design \textbf{Event-Aligned Query Steering (EAQS)} to dynamically align frame-level attention with relevant event prompts for accurate temporal localization.
    
    \item We introduce \textbf{Auto-Balance Strength Solver (ABSS)} to adaptively balance steering strength, maintaining temporal and visual consistency.

\end{itemize}

\begin{figure*}[t]
  \centering
  \includegraphics[width=\linewidth]{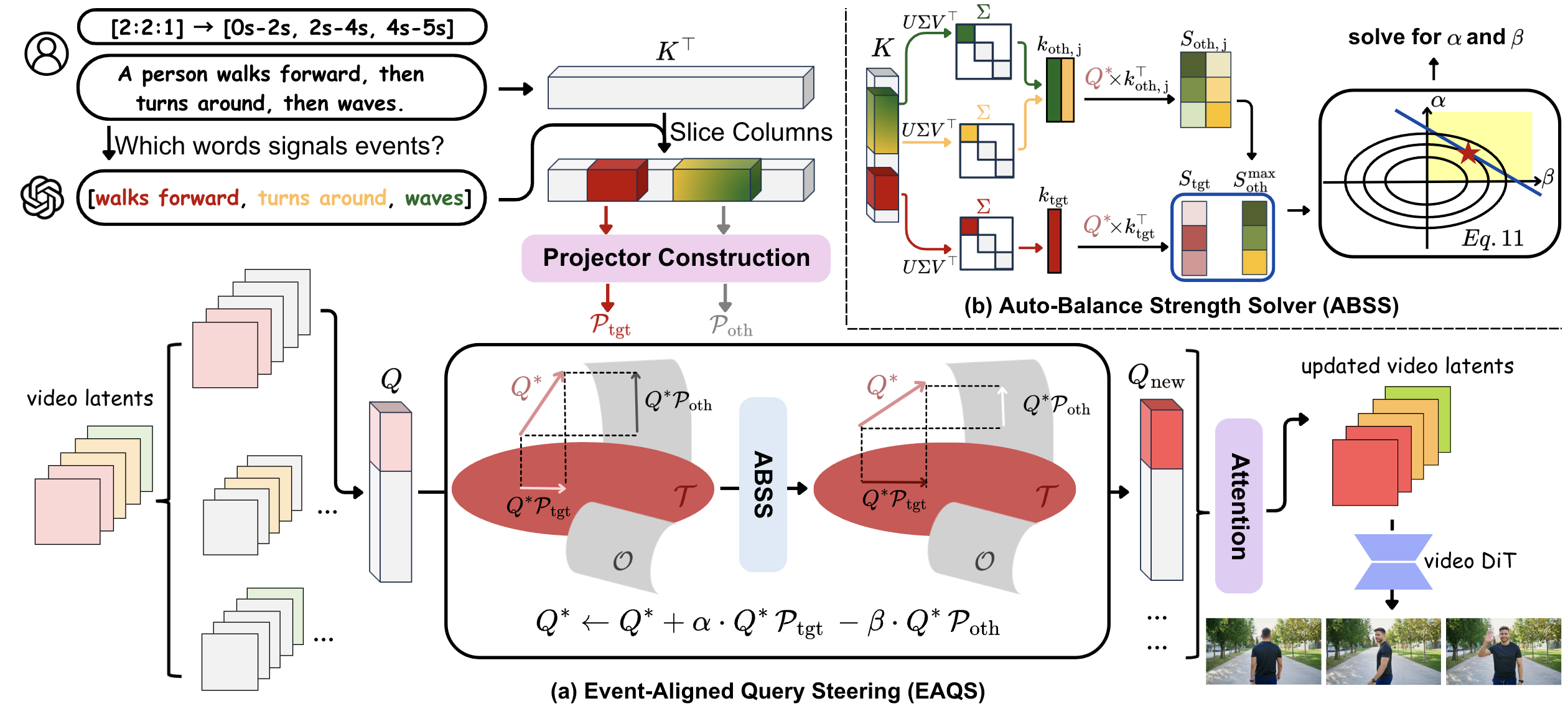}
  \caption{\textbf{Overview of SwitchCraft.} (a) \textbf{EAQS} takes a text prompt and user specified event time spans, identifies anchor tokens for each event, and constructs event specific projectors from their attention keys. It steers video queries toward the target event and away from others in each temporal span. (b) \textbf{ABSS} estimates enhancement and suppression strengths by extracting dominant directions from the event keys and correcting the attention deficit. The updated queries pass through the video diffusion transformer so that each temporal span follows its intended event with smooth transitions.}
  \label{fig:flow}

\end{figure*}

\section{Related Work}
\label{sec:rl}

\textbf{Text-to-video diffusion models.}
T2V diffusion models have advanced rapidly in visual quality, duration and controllability \citep{Wang2025LaVie,Zhang2023Show1,Chen2023VideoCrafter1,Chen2023SEINE,Guo2023AnimateDiff, Harvey2022FDM,Yin2023NUWAXL,Qiu2024FreeNoise}. Early methods adapt pretrained image diffusion models to videos by inserting temporal attention and extending U-Net style denoisers to operate over space and time \citep{Ho2022VDM,He2022LVDM,Blattmann2023AlignYourLatents,Ho2022ImagenVideo,Singer2022MakeAVideo,Wang2025LaVie,Chen2023VideoCrafter1}. More recent architectures replace the U-Net backbone with transformer based diffusion models that treat video as a sequence of space time tokens, which improves scalability to higher resolution, longer duration and more complex motion \citep{BarTal2024Lumiere,Ma2024Latte,Yang2024CogVideoX}. 

Another line of work adopts an autoregressive formulation. Diffusion transformers are adapted into causal generators with key value caching and streaming inference, enabling low latency, frame by frame rollout for real time video generation \citep{yin2025causvid,gao2025ca2vdm,Yang2025LongLive,huang2025selfforcing}. This improves responsiveness but also makes long term stability difficult, since small early errors can accumulate and cause drift. Recent methods address this exposure bias by training on the model’s own streamed rollouts and by using rolling window objectives, cache sharing, and attention anchoring to preserve identity and scene layout over extended sequences \citep{gao2025ca2vdm,liu2025rollingforcing,huang2025selfforcing,cui2025selfforcingpp}. These advances improve duration and streaming quality, but they do not explicitly organize multiple semantically distinct events. Our method addresses this goal directly.

\noindent\textbf{Multi-event video generation.}

Multi-event video generation aims to produce one video with several events in a specified order while preserving scene layout, identities and motion~\citep{Zhu2023MovieFactory,Chen2023SEINE,Long2024VideoStudio,Zhou2024StoryDiffusion,Kara2025ShotAdapter}. A common strategy is sequential stitching. Phenaki~\citep{Villegas2022Phenaki}, Gen-L-Video~\citep{Wang2023GenLVideo} and MEVG~\citep{Oh2024MEVG} generate one segment per prompt, conditioning each new segment on its text and on visual context from previous frames to keep appearance and background consistent. This improves continuity but offers no global view of all events and no explicit control over event duration or boundaries. LongLive~\citep{Yang2025LongLive} extends this paradigm to minute scale streaming videos with cached features, but its causal design blocks access to future events and the accumulated history induces prompt inertia and long range drift. DiTCtrl~\citep{Cai2025DiTCtrl} instead edits attention in a pretrained diffusion transformer to enforce new prompts over time, but this often harms scene consistency and visual quality. Mind the Time~\citep{Wu2025MinT} binds event captions to explicit time spans and fine tunes a diffusion transformer with time aware cross attention, which gives temporally grounded control but demands expensive fine tuning and dense temporal labels. In contrast, our method is training free and provides global control over multiple semantically distinct events while maintaining coherence across the entire video.
\section{Method}

SwitchCraft controls how a pretrained T2V diffusion model attends to multiple events over time. We build on a diffusion transformer backbone and modify its cross attention at inference to align each temporal span with its designated event. Section \ref{prelim} reviews the backbone and its attention mechanism. Section \ref{eaqs} introduces EAQS, which defines an event-specific query update that strengthens alignment with the text features of the target event while suppressing responses to other events. Section \ref{abss} then presents ABSS, which infers the corresponding enhancement and suppression strengths dynamically and supplies them to EAQS. Figure \ref{fig:flow} provides an overview of our pipeline.

\subsection{Preliminaries}
\label{prelim}

\textbf{Text-to-video Diffusion Transformers (DiTs).}
Given a video with $F$ frames and size $H \times W$, a video tokenizer encodes it as 
$z \in \mathbb{R}^{C \times F' \times H' \times W'}$ with temporally and spatially downsampled dimensions. 
The latent tensor is reshaped into a token sequence, and a text conditioned diffusion transformer maps noisy latents $z_t$ toward the clean latents using a velocity prediction under text conditioning $y$:
\begin{equation}
\mathcal{L}
=
\mathbb{E}_{z_t,t,y}\bigl[
\|\,u_\theta(z_t,t,y) - v_t^\star\,\|_2^2
\bigr],
\end{equation}
where $u_\theta$ is the transformer denoiser and $v_t^\star$ is the target velocity. 

\noindent\textbf{Cross attention mechanism.}
T2V DiTs inject prompt information into video features through cross attention. Let \(X \in \mathbb{R}^{S \times d}\) be the video token sequence at the current denoising step and \(Y \in \mathbb{R}^{N \times d}\) the text tokens from the prompt. The layer forms queries from video and keys and values from text with trainable weight matrices \(W_Q,W_K,W_V\),
\begin{equation}
Q = X W_Q \,, \quad
K = Y W_K \,, \quad
V = Y W_V \, ,
\end{equation}
then measures video–text alignment with scaled dot products and injects text guidance back into the video stream,
\begin{equation}
A = \mathrm{softmax}\!\bigl(d_h^{-\tfrac12} Q K^\top\bigr) \,, \quad
\mathrm{out} = A V \, .
\label{attn}
\end{equation}
Here \(d_h\) is the per-head feature dimension and \(\mathrm{out}\) is the cross-attention output that updates the video tokens. In current models this attention is shared across all frames, so a text token that describes a single event can affect every timestep. SwitchCraft instead edits the frame queries within each temporal span, steering attention so that each event primarily governs its designated segment.

\subsection{Event-Aligned Query Steering (EAQS)}
\label{eaqs}

EAQS defines a span-local query update that is controlled by two non-negative steering strengths for enhancement and suppression. In this subsection we treat these strengths as given and focus on how to construct event specific directions from the text keys and how to use them to modify video queries inside the corresponding temporal spans. The procedure that selects the strengths for each span is described in Section~\ref{abss}.

EAQS enforces event-specific temporal alignment in three steps: it decomposes the prompt into distinct events and extracts anchor tokens for each, assigns each event to a dedicated latent temporal window, and updates queries within each window to increase alignment with the target anchors while suppressing competing events.

\noindent\textbf{Event-specific anchor identification.}
We use a large language model such as ChatGPT~\cite{OpenAI2023GPT4} to extract event anchors, defined as a small set of words or short phrases that distinguishes one event description from the others. For example, in scene transitions the anchors focus on setting descriptors such as sunny desert or icy cave, and in multi-event behavior they focus on concise action phrases such as walking forward or reading a book. The instruction prompt that we provide for the language model can be found in the supplementary material. These anchors are then mapped through the backbone tokenizer and recorded as index sets of text tokens for that event.

\noindent\textbf{Temporal window assignment.}
After extracting event phrases, SwitchCraft assigns each event to a temporal span in the latent space. The user may specify relative duration weights $\{w_1, \dots, w_A\}$ for the $A$ events, or choose to default to an equal split. For a video of $F'$ latent frames, the discrete frame count $N_i$ allocated to the $i$-th event is proportionally defined as:
\begin{equation}
N_i \approx F' \cdot \frac{w_i}{\sum_{j=1}^A w_j}.
\end{equation}
To ensure exactly $F'$ frames are assigned without gaps, we round these targets to integers and distribute any truncation remainder to the events with the largest fractional parts. The $i$-th event is then mapped to the contiguous, half-open frame index interval $[\sum_{j=1}^{i-1} N_j, \sum_{j=1}^i N_j)$. 

\noindent\textbf{Query Steering.}  
As defined in Equation \eqref{attn}, cross attention governs how each frame draws information from the text tokens. To strengthen alignment with the text, we aim to increase the values of \(QK^{\top}\) associated with the target tokens within the event span and to reduce the entries associated with tokens from other events. Since keys and values are shared across frames, modifying them would propagate effects to all frames and can degrade frames not intended to change. We therefore edit only the queries, which are frame-specific and can steer attention only within their assigned event span. 

We steer queries at the level of individual heads to obtain fine grained temporal control. Inside one cross attention block and for a single head, let \(K \in \mathbb{R}^{L_k \times D}\) be the matrix of text keys and let \(Q^{*} \in \mathbb{R}^{R \times D}\) be the slice of query rows for the latent frames assigned to the current event, with \(L_k\) the number of text tokens, \(D\) the feature dimension, and \(R\) the number of rows corresponding to the video tokens in this span. We form $K_{\text{tgt}}$ by gathering from $K$ the rows corresponding to the target event’s anchor tokens, and form $K_{\text{oth}}$ by gathering the rows corresponding to anchor tokens from the remaining events. We want every row of $Q^*$ to produce higher dot products with columns from its corresponding $K^\top_{\text{tgt}}$ and lower dot products with columns from $K^\top_{\text{oth}}$. The most direct way to achieve this is to decompose queries into components relative to the spans of \(K_{\text{tgt}}\) and \(K_{\text{oth}}\), then enhance the component that lies in the span of \(K_{\text{tgt}}\) and suppress the component that lies in the span of \(K_{\text{oth}}\). 

To carry out this decomposition we need an operator that returns, for any query, its component within a specified key span. A naive attempt is to compute \(Q^{*}\cdot K^{\top}\), but this reports token-level instead of event-level alignment, and lives in \(\mathbb{R}^{R \times L_k}\) which is not the same space as \(Q^{*} \in \mathbb{R}^{R \times D}\). We therefore construct right projectors $P_{\text{tgt}}$ and $P_{\text{oth}}$ from $K_{\text{tgt}}$ and $K_{\text{oth}}$, that can project $Q^*$ to their key spans defined by
\begin{equation}
P_{\text{tgt}} \;=\; \mathcal{P}[K_{\text{tgt}}] \;=\; K_{\text{tgt}}^{\top}\bigl(K_{\text{tgt}}K_{\text{tgt}}^{\top}+\epsilon I\bigr)^{-1}K_{\text{tgt}},
\end{equation}
with small ridge \(\epsilon>0\) and identity \(I\), and analogously \(P_{\text{oth}}=\mathcal{P}[K_{\text{oth}}]\). We view these spans as two subspaces in the $D$-dimensional key space: $\mathcal{T}$ for the target event, spanned by $K_{\text{tgt}}$, and $\mathcal{O}$ for all competing events, spanned by $K_{\text{oth}}$ (see Figure~\ref{fig:flow}). The projectors $P_{\text{tgt}}$ and $P_{\text{oth}}$ therefore map $Q^{*}$ onto $\mathcal{T}$ and $\mathcal{O}$, respectively.

To reinforce the target event within its temporal span and reduce the influence of unrelated events, we modulate the queries with nonnegative strengths \(\alpha\) and \(\beta\):
\begin{equation}
Q^* \leftarrow Q^* + \alpha \cdot \, Q^* P_{\text{tgt}} - \beta \cdot \, Q^* P_{\text{oth}} .
\label{q}
\end{equation}
The first term boosts the component of $Q^*$ that lies in the span of $\mathcal{T}$, which increases alignment with the keys of the intended event. The second term subtracts the component of $Q^*$ that lies in the span of $\mathcal{O}$, which weakens alignment with keys that correspond to other events and reduces leakage. After this edit we apply a row-wise renormalization to stabilize attention magnitudes. This update shifts the queries of frames in the span toward the target event, so the generated content follows the intended instruction while the influence of unrelated events is suppressed.

Notably, we apply the edit only in the early denoising steps and in earlier blocks. This choice is motivated by how diffusion transformers organize generation over time and depth. Early denoising steps and early transformer blocks establish scene layout and large scale motion, while later steps and later blocks primarily refine texture, identity, and appearance detail \cite{Ma2024Latte,Yang2024CogVideoX,Lv2025DCM}. Steering in this early stage is therefore sufficient to enforce when each event should occur. After this point the unmodified base model can complete high frequency detail without further intervention. Additionally, because the update acts on the queries rather than on the attention weights after the softmax, it preserves the structure learned by the pretrained model and avoids abrupt changes at event boundaries.

\subsection{Auto-Balance Strength Solver (ABSS)}
\label{abss}
To adapt the steering strength to different prompts and scenes, we introduce ABSS, which robustly determines the enhancement \(\alpha\) and suppression \(\beta\) used in the update in Equation \ref{q}. The goal is to ensure that the current event dominates attention within its assigned temporal span while avoiding excessive edits that would distort appearance or destabilize motion. ABSS operates at inference and measures, on the frames of the active span, the relative alignment of the current queries with the target span \(K_{\text{tgt}}\) against the competing span of \(K_{\text{oth}}\). It then solves a two-variable constrained quadratic program to select nonnegative \(\alpha\) and \(\beta\).

We aim to correct the deficit in each event span. A direct comparison based on \(\Delta = Q^{*}K_{\text{oth}}^{\top} - Q^{*}K_{\text{tgt}}^{\top}\) and optimizing from \(\Delta\) would be high dimensional and expensive to evaluate, sensitive to token count and order, and does not yield a stable feature space edit. We therefore compress the keys of each event into one dominant direction by applying singular value decomposition to the normalized key rows. Denote these unit directions by $k_{\text{tgt}}$ for the current event and $k_{\text{oth},j}$ for each competitor event $j = 1,\dots,J$. This yields one representative direction per event that is robust to the number and location of anchor tokens. 
Let $k_{\text{oth}} = [\,k_{\text{oth},1}\;\cdots\;k_{\text{oth},J}\,] \in \mathbb{R}^{D\times J}$ denote the matrix whose columns are the competitor directions $k_{\text{oth},j}$. We then define row-wise alignment scores $S_{\text{tgt}}\in\mathbb{R}^{R}$ and $S_{\text{oth}}\in \mathbb{R}^{R\times J}$ for $Q^* \in \mathbb{R}^{R \times D}$ by
\begin{equation}
S_{\text{tgt}} = Q^* \, k_{\text{tgt}}, \quad
S_{\text{oth}} = Q^* \, k_{\text{oth}} \,.
\end{equation}

For every row, take the dominant competitor
\(S_{\text{oth}}^{\max} = \max_j S_{\text{oth},j}\) and define the margin deficit
\begin{equation}
d = S_{\text{oth}}^{\max} - S_{\text{tgt}} + \varepsilon \, .
\end{equation}

\begin{figure*}[t]
  \centering
  \includegraphics[width=\linewidth]{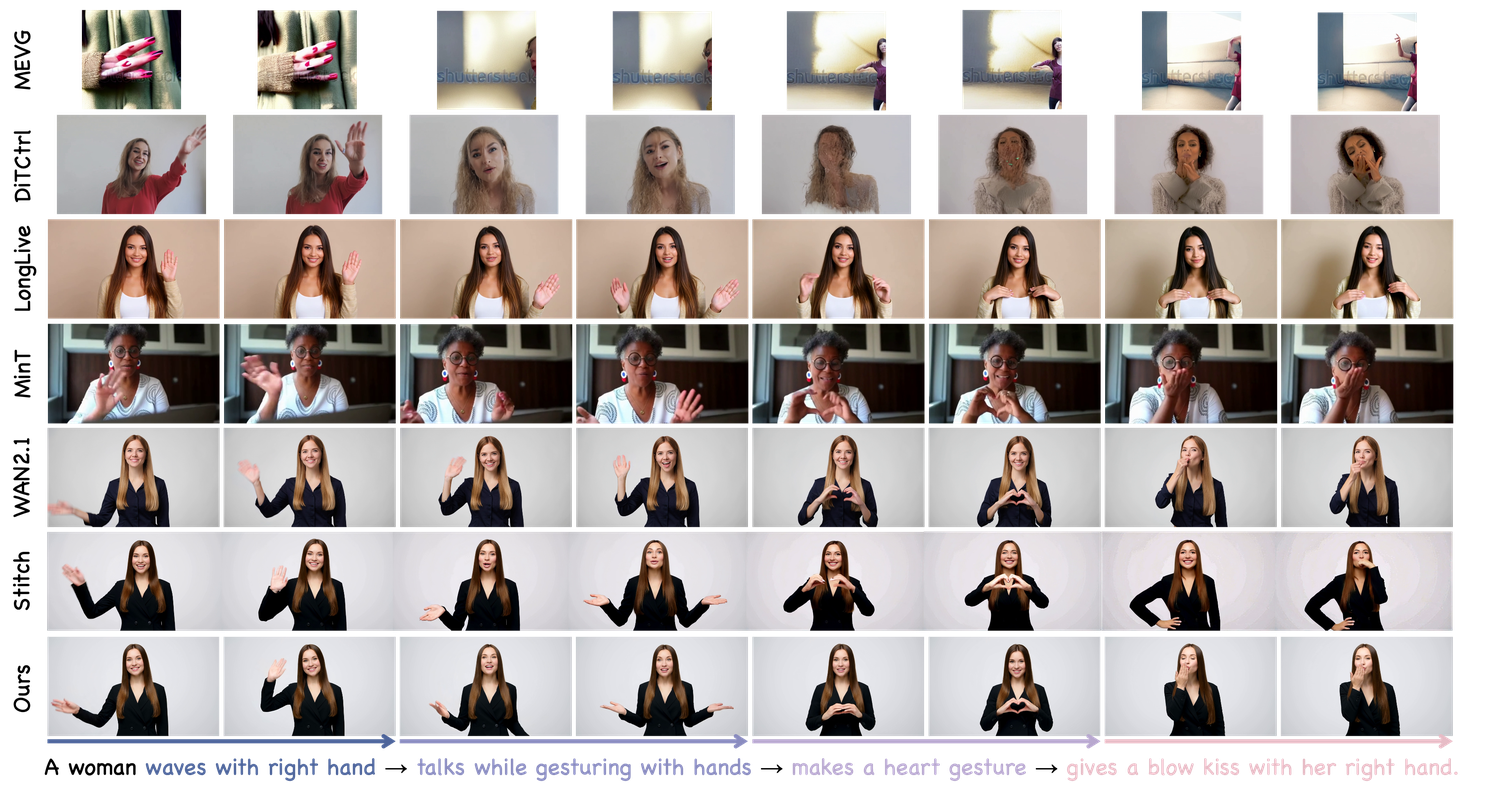}
  \caption{\textbf{Qualitative comparison.} SwitchCraft executes all events in the intended order, prevents leakage and omission, and maintains subject and scene consistency. Baselines show omissions, drift across spans, or progressive quality decay.}
  \label{fig:comparison_method}
\end{figure*}

To solve for the steering strengths $\alpha$ and $\beta$ we formulate a convex minimization problem that balances two requirements. The first is to make the target event dominate within the current span by raising the margin against competing events. The second is to prevent oversteering by discouraging unnecessarily large values. For the first requirement, to compensate for the deficit \(d\) we require that the combined effect of enhancement and suppression satisfies: 
\begin{equation}
\alpha \, S_{\text{tgt}} \;+\; \beta\,S_{\text{oth}}^{\max} \;\ge\; d .
\end{equation}

Additionally, to keep the edit conservative and protect appearance, we introduce a resistance matrix that measures how expensive it is to move along each steering direction. We penalize large strengths using a diagonal matrix $M$ whose entries reflect how sensitive the margin is to changes in $\alpha$ and $\beta$ over the span:
\begin{equation}
M =
\begin{bmatrix}
\|S_{\text{tgt}}\|_2^2 & 0 \\
0 & \bigl\|S_{\text{oth}}^{\max}\bigr\|_2^2
\end{bmatrix}.
\end{equation}

We penalize large strengths using a quadratic term \(x^\top M x\) with \(x = [\,\alpha\;\;\beta\,]^\top\). The diagonal entries of \(M\) quantify how strongly changes in \(\alpha\) and \(\beta\) affect the margin over the span, so directions that already induce large margin changes receive a higher cost. This yields scale-aware damping, stabilizes the update, and avoids pushing too far.

Finally, with \(C=[\,S_{\text{tgt}}\;\; S_{\text{oth}}^{\max}]\), the convex minimization problem is formulated by:
\begin{equation}
\min_{x\ge 0}\;
\frac{1}{2}\, x^{\top} M x
\;+\;
\frac{1}{2}\, \bigl\|\max\!\bigl(0,\, d - Cx\bigr)\bigr\|_{2}^{2}.
\end{equation}

The 
second term is inactive when \(d-Cx\le 0\), in which case the minimum is \(x=0\). When it is active the unique stationary point satisfies
\begin{equation}
\bigl(M + C^{\top} C\bigr)\,x \;=\; C^{\top} d,
\quad
x \leftarrow \max(x,0).
\label{eq to prove}
\end{equation}
The full derivation can be found in the supplementary material.
The final \(\alpha\) and \(\beta\) are applied uniformly within the current event span.

\definecolor{AlignColor}{RGB}{255,249,219}   
\definecolor{SmoothColor}{RGB}{242,234,255}  
\definecolor{VisualColor}{RGB}{232,240,254}  

\section{Experiments}
\begin{table*}[t]
\centering
\setlength{\tabcolsep}{3pt}
\caption{\textbf{Quantitative comparison and ablation study.} SwitchCraft improves text alignment and preserves temporal smoothness and visual quality across CLIP \cite{Radford2021CLIP,Hessel2021CLIPScore}, VideoScore2 \cite{He2025VideoScore2}, and VBench \cite{Huang2023VBench}. Best scores are in \textbf{bold} and second best are \underline{underlined}. Columns highlighted in \colorbox{AlignColor}{yellow} correspond to text alignment metrics,
\colorbox{SmoothColor}{purple} to temporal smoothness metrics,
and \colorbox{VisualColor}{blue} to visual quality and consistency metrics.
Removing components in the ablations generally reduces text alignment,
while the other metrics remain close to the Wan 2.1 ~\cite{wan2025} backbone.}

\label{tab:quant}
\resizebox{\textwidth}{!}{
\begin{tabular}{l cc ccc ccccc}
\toprule
 & \multicolumn{2}{c}{CLIP \cite{Radford2021CLIP,Hessel2021CLIPScore}} & \multicolumn{3}{c}{VideoScore2 \cite{He2025VideoScore2}} & \multicolumn{5}{c}{VBench \cite{Huang2023VBench}} \\
\cmidrule(lr){2-3}\cmidrule(lr){4-6}\cmidrule(lr){7-11}
Method
& \cellcolor{AlignColor} CLIP-T $\uparrow$ 
& \cellcolor{SmoothColor} CLIP-F $\uparrow$
& \cellcolor{VisualColor} Visual quality $\uparrow$ 
& \cellcolor{AlignColor} T2V alignment $\uparrow$ 
& \cellcolor{VisualColor} Phys. cons. $\uparrow$
& \cellcolor{SmoothColor} Motion smoothness $\uparrow$ 
& \cellcolor{VisualColor} Sub cons. $\uparrow$ 
& \cellcolor{VisualColor} Back cons. $\uparrow$ 
& \cellcolor{VisualColor} Aesthetic quality $\uparrow$ 
& \cellcolor{VisualColor} Imaging quality $\uparrow$ \\
\midrule
MEVG \cite{Oh2024MEVG}    & 0.244 & 0.915 & 2.13 & 2.33 & 1.73 & 0.953 & 0.701 & 0.841 & 0.346 & 0.525 \\
DiTCtrl \cite{Cai2025DiTCtrl}  & 0.246 & 0.959 & 3.20 & 3.27 & 2.93 & 0.981 & 0.764 & 0.876 & 0.511 & 0.702 \\
LongLive ~\cite{Yang2025LongLive} & 0.252 & \textbf{0.984} & 4.27 & 3.13 & 3.97 & 0.984 & 0.898 & 0.908 & 0.627 & 0.725 \\
Wan2.1 ~\cite{wan2025}     & 0.256 & \underline{0.980} & \underline{4.30} & 3.47 & \underline{4.12} & \underline{0.987} & \textbf{0.947} & \textbf{0.924} & \underline{0.645} & \underline{0.738} \\
Stitch   & \underline{0.257} & 0.963 & 3.73 & \underline{3.67} & 3.80 & 0.983 & 0.926 & 0.910 & 0.608 & 0.711 \\
\cmidrule(lr){1-11}
\rowcolor{gray!10}
\textbf{Ours} &
\textbf{0.275} & \underline{0.980} & \textbf{4.33} & \textbf{4.30} & \textbf{4.13} & \textbf{0.989} & \underline{0.945} & \underline{0.921} & \textbf{0.648} & \textbf{0.741} \\
\cmidrule(lr){1-11}
Random strength & 0.253 & 0.974 & 4.15 & 3.62 & 3.98 & 0.987 & 0.939 & 0.915 & 0.637 & 0.734 \\
Fixed strength & 0.264 & 0.967 & 3.97 & 3.75 & 3.95 & 0.985 & 0.934 & 0.907 & 0.631 & 0.715 \\
\textit{w/o} SVD & 0.255 & 0.978 & 4.30 & 3.67 & 4.08 & 0.988 & 0.943 & 0.918 & 0.643 & 0.734 \\
\textit{w/o} enhance & 0.262 & 0.980 & 4.35 & 3.78 & 4.13 & 0.989 & 0.945 & 0.922 & 0.646 & 0.739 \\
\textit{w/o} suppress & 0.261 & 0.978 & 4.28 & 3.73 & 4.05 & 0.986 & 0.942 & 0.920 & 0.642 & 0.736 \\
\bottomrule
\end{tabular}}
\end{table*}

\subsection{Experimental Setup}
We implement SwitchCraft on the Wan\,2.1 T2V 14B backbone~\cite{wan2025} and generate videos with a standard resolution of $832 \times 480$ pixels and length of 5 seconds on one NVIDIA A100 GPU. Sampling uses 50 denoising steps and the query steering is enabled only in the first 20 blocks of the first 20 steps to shape layout and motion, and edits are disabled for the remaining blocks and steps so the base model refines appearance. All other sampling parameters follow the Wan 2.1 default configuration.

We evaluate on 60 multi-event prompts that cover action switches and scene transitions. Prompts are drawn from prior work and from an automatically generated set produced by a large language model~\cite{OpenAI2023GPT4}, spanning two to four events per video. We compare SwitchCraft to baselines of training-based temporal controllers MinT~\cite{Wu2025MinT} and LongLive~\cite{Yang2025LongLive}, training-free multi-event controller MEVG~\cite{Oh2024MEVG} and DiTCtrl~\cite{Cai2025DiTCtrl}, and the Wan\,2.1 base model~\cite{wan2025}. We also design a stitching baseline \textit{Stitch} that generates consecutive segments with the Wan\,2.1 image-to-video (I2V) model using the last frame of the previous segment as the starting frame. Note that MinT~\cite{Wu2025MinT} does not release its base model, trained weights, or training data, so we compare to its public samples for qualitative analysis and human evaluation only. For all baselines, we follow their default experimental setups.

We evaluate text alignment, temporal smoothness, and visual quality and consistency. Text alignment is measured by per-frame CLIP-T~\cite{Radford2021CLIP,Hessel2021CLIPScore} and by the text-to-video alignment score in VideoScore2~\cite{He2025VideoScore2}. 
Temporal smoothness is measured by CLIP-F, which computes cosine similarity between CLIP image embeddings of adjacent frames~\cite{Radford2021CLIP,Hessel2021CLIPScore}, and by the motion smoothness index in VBench~\cite{Huang2023VBench}. 
Visual quality is assessed using the visual quality and physical consistency scores in VideoScore2~\cite{He2025VideoScore2}, and the subject consistency, background consistency, aesthetic quality (perceived visual appeal), and imaging quality (technical fidelity) scores in VBench \cite{Huang2023VBench}.

\subsection{Qualitative Results}

\begin{figure}[t]
  \centering
  \includegraphics[width=\linewidth]{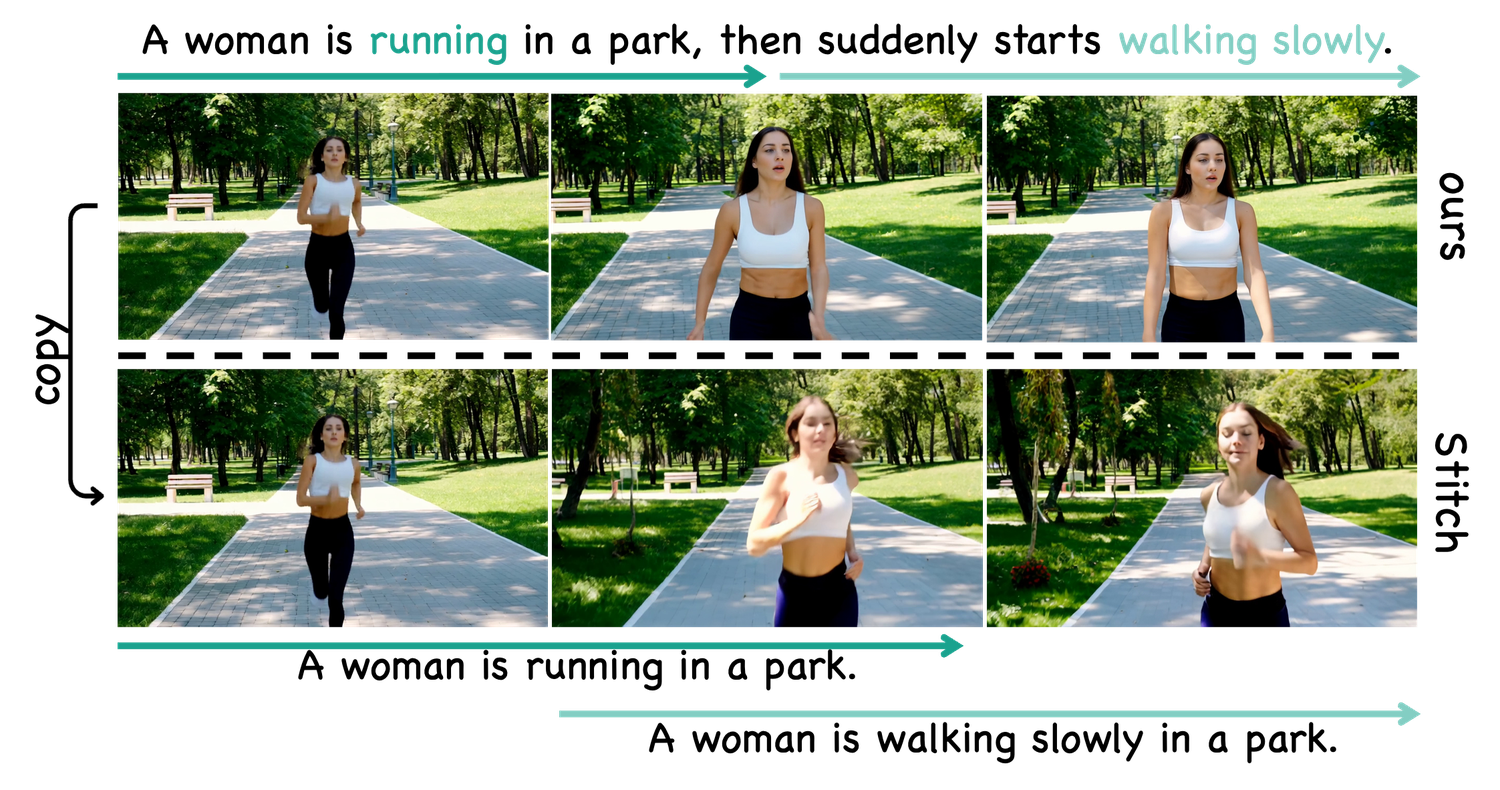}
  \caption{\textbf{Failure of \textit{Stitch}.} Segment-wise generation inherits bias from the last frame of the previous segment and propagates motion and layout, causing action bleed and unstable handoffs.}
  \vspace{-10pt}
  \label{fig:stitch}
\end{figure}

\textbf{Qualitative comparison.} Figure~\ref{fig:comparison_method} compares SwitchCraft with strong baselines on a four-event prompt from the released work of MinT~\cite{Wu2025MinT}. MEVG~\cite{Oh2024MEVG} exhibits low visual fidelity due to its earlier backbone and converts each event into a new viewpoint, breaking identity and background continuity. DiTCtrl~\cite{Cai2025DiTCtrl} shows a similar structural weakness: transitions are visually smooth, yet the global scene and subject drift from event to event. LongLive~\cite{Yang2025LongLive} produces only a small subset of the required events, actions stagnate and frame quality degrades over time. MinT~\cite{Wu2025MinT} generates coherent actions aligned with the text but leaks activity across spans, as the subject continues talking outside the designated segment. The base Wan 2.1~\cite{wan2025} preserves appearance but misses later events and blurs boundaries, with the first action persisting into subsequent segments. Stitch baseline yields seemingly reasonable sequences but is highly sensitive to the previous key frame and inherits its motion bias, causing action bleed and occasional discontinuities, as corroborated by the failure in Figure~\ref{fig:stitch}. 
In contrast, SwitchCraft maintains subject and scene, executes all events in the correct order, avoids leakage and omission, and sustains visual quality over time. More qualitative results can be found in the supplementary material.

\begin{figure}[t]
  \centering
  \includegraphics[width=\linewidth]{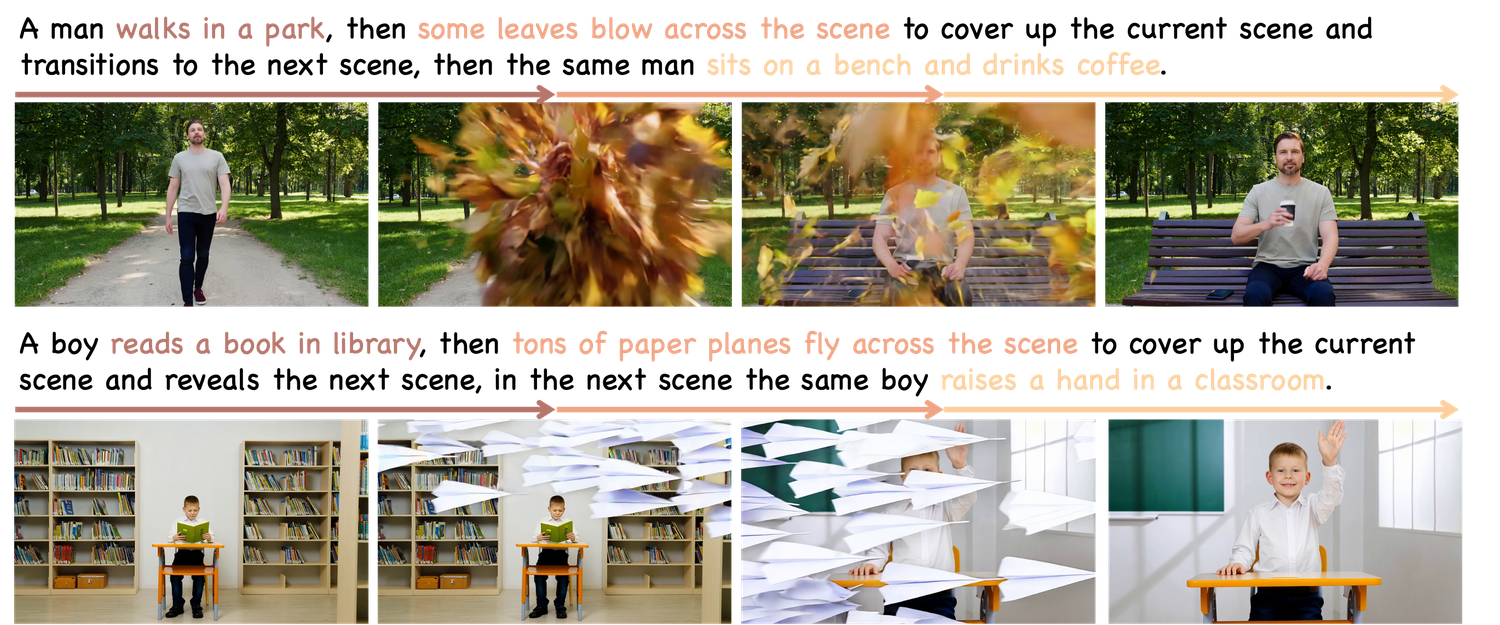}
  \caption{\textbf{More applications.} SwitchCraft generates creative occluding transitions between scenes, preventing event bleeding while preserving identity and global context.}
  \label{fig:trans}
\end{figure}

\noindent\textbf{More applications.} As shown in Figure~\ref{fig:trans}, SwitchCraft can produce creative transition effects when the middle segment of the script describes a visual occluder that gradually masks the current scene and reveals the next one.  
Because the edit acts only on video queries to steer to specific events, subject identity and global context can propagate across segments while actions from adjacent segments do not bleed into the transition.

\subsection{Quantitative Results}
\textbf{Quantitative comparison.} We conduct a quantitative comparison between SwitchCraft and other state-of-the-art methods. 
Our experimental results in Table \ref{tab:quant} demonstrate the superior performance of our method that significantly enhances text alignment without degrading motion smoothness and visual quality. Notably, our method does not achieve the highest score in CLIP-F because this metric rewards very similar neighboring frames, so larger pose changes or fast actions reduce the score even when transitions are smooth and realistic. Our scores in subject and background consistency are slightly lower than that of Wan2.1 base model \cite{wan2025} because VBench \cite{Huang2023VBench} measures these consistencies by feature similarity across frames. SwitchCraft emphasizes the requested events and reveals new viewpoints and regions, which may lower feature similarity even when identity and scene remain stable to human observers.

\begin{table}[t]
\centering
\setlength{\tabcolsep}{4pt}
\caption{\textbf{Human evaluation.} Users scored no omission, no leakage, transition smoothness, and visual quality on a five point scale. SwitchCraft attains the best overall performance.}
\label{tab:userstudy}
\resizebox{\columnwidth}{!}{
\begin{tabular}{lcccc}
\toprule
Method & No omission $\uparrow$ & No leakage $\uparrow$ & Smooth transition $\uparrow$ & Visual quality $\uparrow$ \\
\midrule
MEVG \cite{Oh2024MEVG}    & 1.41 & 1.38 & 1.38 & 1.28 \\
DiTCtrl \cite{Cai2025DiTCtrl} & 1.66 & 1.48 & 1.48 & 1.59 \\
LongLive \cite{Yang2025LongLive} & 2.07 & 2.72 & 2.97 & 3.52 \\
MinT \cite{Wu2025MinT}     & \textbf{4.31} & \underline{3.69} & 3.76 & 3.83 \\
Wan 2.1 \cite{wan2025}     & 3.17 & 3.38 & \underline{3.79} & \underline{3.93} \\
Stitch   & 2.62 & 2.07 & 2.14 & 2.45 \\
\cmidrule(lr){1-5}
\rowcolor{gray!10}
\textbf{Ours} & \underline{4.21} & \textbf{4.04} & \textbf{3.93} & \textbf{4.24} \\
\bottomrule
\end{tabular}}
\end{table}

\noindent\textbf{Human evaluation.} To complement automatic metrics and assess human preferences, we conduct a user study across SwitchCraft and all baselines. We invited 29 users and use a five point scale where 5 is best. The participants are asked to score four aspects: no omission meaning all requested events appear, no leakage meaning events do not appear outside their spans, smooth event transition, and visual quality. As shown in Table \ref{tab:userstudy}, SwitchCraft achieves the highest mean scores on no leakage, transition smoothness, and visual quality, and is close to the training-based method MinT \cite{Wu2025MinT} on no omission. These results show that our method delivers the intended event sequence with fewer side effects and high perceived quality.

\subsection{Ablation Study}
\begin{figure}[t]
  \centering
  \includegraphics[width=\linewidth]{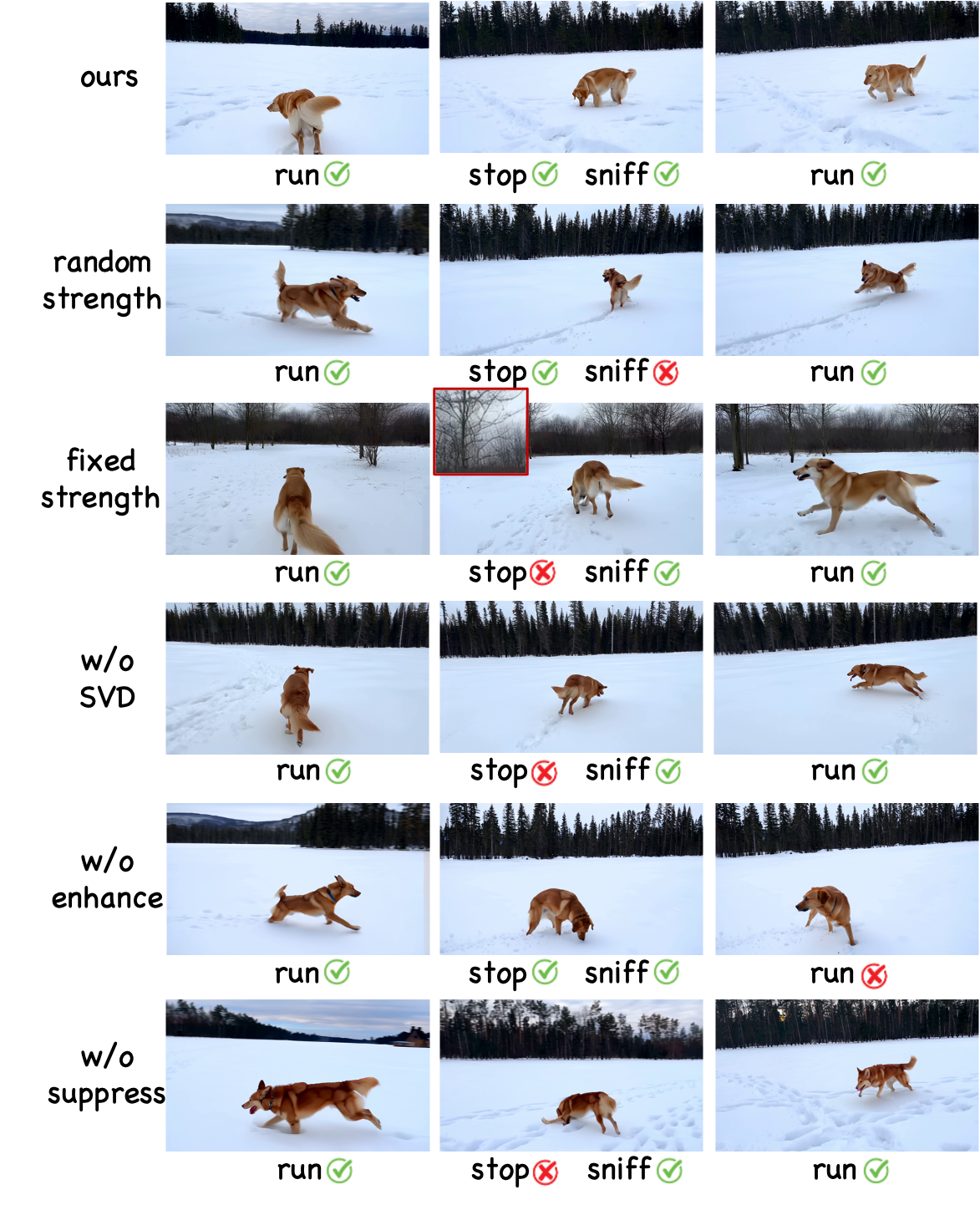}
  \caption{\textbf{Ablation study of key components.} Prompt: \textit{On a snowy plain, a dog is running forward, then suddenly stops to sniff the ground, then continues running.} The full method yields complete sequences with stable quality. Removing parts leads to event leakage, omission, and degraded visual quality.}
  \label{fig:ablation}
\end{figure}

We conduct ablations to assess the contribution of each component of SwitchCraft. As shown in Table~\ref{tab:quant} and Figure~\ref{fig:ablation}, the full model executes all events in the correct spans while maintaining high visual quality and consistency. 

\noindent\textbf{Effectiveness of ABSS.} Removing ABSS and sampling random strengths from 0 to 1 causes missed or delayed events, and fixing the strength to 1 leads to over-steering and degraded appearance. Removing the SVD weakens the separation between events in key space, which lowers text alignment while keeping frame quality close to the backbone. 

\noindent\textbf{Effectiveness of EAQS.} In EAQS, the suppression-only variant (no enhancement) reduces leakage of events but does not actively drive queries toward the next event, so later events often reduce and disappear. Enhancement-only (no suppression) fails to isolate spans when competitors are strong, so dominant actions persist and blend with the target event, introducing artifacts and reducing both alignment and perceived quality. These results show that all key components are necessary to realize the intended multi-event sequence while preserving visual quality.

\section{Conclusion}

In this work, we presented SwitchCraft, a training-free framework for multi-event video generation. Our Event-Aligned Query Steering modifies video queries within each temporal span so that they preferentially attend to the target event and are repelled from competing events, while the Auto-Balance Strength Solver dynamically selects enhancement and suppression strengths, avoiding under- or over-steering. Experiments across automatic metrics and human evaluations show that SwitchCraft improves text alignment, temporal smoothness, and visual quality over baselines.

\section*{Acknowledgement}

This work was supported by the National Natural Science Foundation of China (No. 6250070674) and the Zhejiang Leading Innovative and Entrepreneur Team Introduction Program (2024R01007).

{\small
\bibliographystyle{ieeenat_fullname}
\bibliography{main}
}
\clearpage
\setcounter{page}{1}

\appendix

\paragraph{Supplementary Overview.}
This supplementary material presents additional details and analyses
supporting the main paper.
Section~\ref{app:implementation} reports implementation details and hyperparameters of the
Wan~2.1 T2V 14B backbone\cite{wan2025} used by SwitchCraft.
Section~\ref{app:abss_proof} provides the derivation of the closed-form update in ABSS.
Section~\ref{app:llm_prompts} specifies the prompts used to query large language models for
event-specific anchors and for generating multi-event evaluation
prompts.
Section~\ref{sec:limitations} discusses the main limitations of our approach.
Section~\ref{sec:inference_time} reports inference time for different numbers of events.
Section~\ref{app:occluding_transition} analyzes why SwitchCraft can realize creative occluding
transitions more reliably than prior methods.
Section~\ref{sec:add_qualitative} presents additional qualitative comparisons and SwitchCraft
results.

\section{Implementation Details}
\label{app:implementation}

We build SwitchCraft on the Wan 2.1 T2V 14B backbone \cite{wan2025}. In addition to the hyperparameters noted in the main paper, others are shown in Table \ref{tab:wan_hparams}.

\section{Proof of Equation 12 in ABSS}
\label{app:abss_proof}

Recall the convex objective used in ABSS that minimizes the function
\begin{equation}
  \frac{1}{2} x^{\top} M x
  +
  \frac{1}{2} \left\|\max\!\bigl(0,\, d - Cx\bigr)\right\|_2^2
\label{eq:abss_objective}
\end{equation}
subject to the constraint \(x \ge 0\).
Here the first term is a quadratic regularizer on \(x\), and the second term
is a squared hinge loss that penalizes violations of the constraint
\(Cx \ge d\).

To derive the update used in ABSS, we study the first order optimality
condition of \eqref{eq:abss_objective} on the region where the hinge is
active, that is, for components with \(d - Cx > 0\).
On this region the hinge simplifies to
\(\max(0, d - Cx) = d - Cx\),
and the objective reduces to the smooth quadratic
\begin{equation}
  \frac{1}{2} x^{\top} M x
  +
  \frac{1}{2} \,\bigl\|d - Cx\bigr\|_2^2 .
\label{eq:abss_quadratic}
\end{equation}

We now compute the gradient of \eqref{eq:abss_quadratic} with respect to \(x\).
For the first term, set
\(
  f(x) = \tfrac{1}{2} x^{\top} M x
\)
and write
\[
  x^{\top} M x
  = \sum_{i=1}^{n} \sum_{j=1}^{n} M_{ij} x_i x_j .
\]
Taking the partial derivative with respect to \(x_k\) and using the product
rule gives
\[
  \frac{\partial f}{\partial x_k}
  =
  \frac{1}{2} \sum_{j} M_{kj} x_j
  +
  \frac{1}{2} \sum_{i} M_{ik} x_i
  =
  \frac{1}{2} \bigl[(Mx)_k + (M^{\top} x)_k\bigr] .
\]
Stacking the components yields
\[
  \nabla_x f(x)
  =
  \frac{1}{2} (M + M^{\top}) x .
\]
In our setting \(M\) is symmetric, so \(M = M^{\top}\) and therefore
\begin{equation}
  \nabla_x \left( \frac{1}{2} x^{\top} M x \right)
  = M x .
\label{eq:grad_M}
\end{equation}

For the second term we apply the chain rule:
\begin{equation}
  \nabla_x
  \left(
    \frac{1}{2} \|d - Cx\|_2^2
  \right)
  =
  - C^{\top} (d - Cx)
  =
  C^{\top} C x - C^{\top} d .
\label{eq:grad_hinge}
\end{equation}
Combining \eqref{eq:grad_M} and \eqref{eq:grad_hinge}, the gradient of the
objective in \eqref{eq:abss_quadratic} becomes
\begin{equation}
  Mx + C^{\top} C x - C^{\top} d .
\label{eq:grad_full}
\end{equation}

Then any stationary point of this quadratic problem satisfies
\begin{equation}
  \nabla_x
  \left(
    \frac{1}{2} x^{\top} M x
    +
    \frac{1}{2} \,\bigl\|d - Cx\bigr\|_2^2
  \right)
  = 0 .
\end{equation}
Using \eqref{eq:grad_full}, this condition yields
\[
  Mx + C^{\top} C x - C^{\top} d = 0
  \quad\Longrightarrow\quad
  (M + C^{\top} C) x = C^{\top} d .
\]
which is the linear system that ABSS solves at each iteration.

Additionally, ABSS enforces the constraint
\(x \ge 0\) by
\begin{equation}
  x \leftarrow \max(x, 0) ,
\end{equation}
which recovers Equation 12 in the main paper.

\begin{table}[t]
  \centering
  \caption{Hyperparameters for the Wan 2.1 T2V 14B \cite{wan2025} backbone used by
  SwitchCraft.}
  \label{tab:wan_hparams}
  \vspace{2pt}
  \begin{tabular}{l c}
    \toprule
    Hyperparameter        & Value \\
    \midrule
    DiT blocks            & \(40\) \\
    Attention heads       & \(40\) \\
    Video frames   & \(81\) \\
    Sampler               & UniPC \\
    Guidance scale        & \(5.0\) \\
    \bottomrule
  \end{tabular}
\end{table}

\label{sec:rationale}

\begin{figure}[t]
  \centering
  \includegraphics[width=\linewidth]{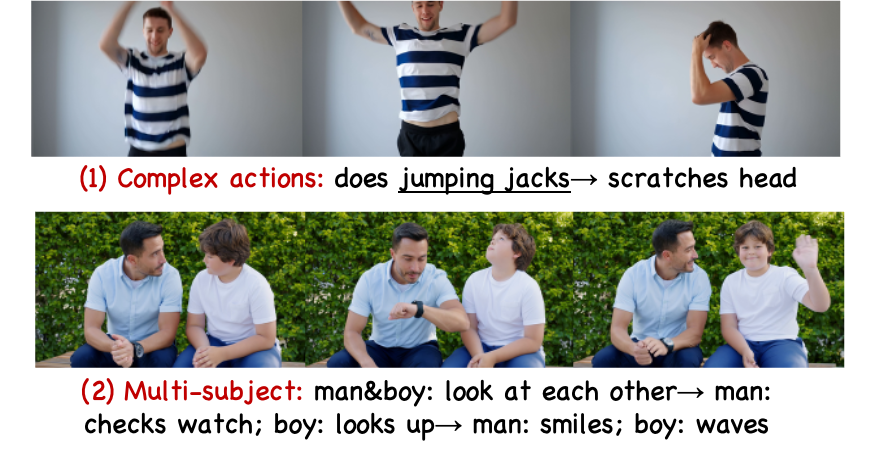}
  \caption{\textbf{Limitations.} (1) Complex actions: When the underlying architecture struggles with specific movements, the output defaults to basic approximations. (2) Multi-subject: Our method does not have spatial constraints for solving identity-action entanglement between characters.}
  \label{fig:limit}
\end{figure}

\section{Prompts for Large Language Models}
\label{app:llm_prompts}

In this work we utilize Large Language Models (LLMs) in two places:
to identify event-specific anchor phrases for Event-Aligned Query
Steering (EAQS), and to generate multi-event prompts for our
evaluation.

\noindent\textbf{Event specific anchor identification.}
Given a multi-event video description, we ask the LLM to extract a
small set of anchor phrases for each event. These anchors are used to
select event specific text tokens. The instruction prompt is:

\begin{quote}
\small\ttfamily
You are an assistant that analyzes a single text prompt describing a
video with multiple temporally ordered events.\\[0.5ex]
Your goal is to identify, for each event, a set of anchor
phrases that clearly distinguish this event from the others. Anchors
should be short noun phrases or verb phrases taken directly from the
prompt, such as setting descriptors like ``sunny desert'' or ``icy
cave'' or concise action phrases like ``walking forward'' or
``reading a book''.\\[0.5ex]
Requirements:\\
1.~Do NOT invent new events. Only use events that are explicitly
   described in the input prompt.\\
2.~Every anchor phrase must be a substring of the original prompt.\\
3.~Omit the shared subject and transitional words. Keep the full remaining verb phrase that describes what is happening in that specific event.\\[0.5ex]
Input format: I will give you one prompt that may contain multiple
events in temporal order.\\
Output format: List all anchor phrases you extract for this prompt on a
single line, separated by commas, with no additional explanations.\\[0.5ex]
Now analyze the following prompt and return the anchors in the exact
format above.
\end{quote}

\noindent\textbf{Multi-event prompt generation.}
We also use an LLM to construct multi-event prompts for our
experiments. The instruction prompt is:

\begin{quote}
\small\ttfamily
You are designing text prompts for a text to video diffusion model.\\[0.5ex]
Your task is to generate natural language prompts that describe
one video containing multiple temporally ordered events that can plausibly occur within a single short continuous clip.\\[0.5ex]
General requirements:\\
1.~Each prompt must describe between two and four distinct events that
   happen in sequence.\\
2.~Use everyday subjects such as people, animals, or common objects in
   realistic scenes.\\
3.~Cover two types of changes: action switches and scene
   transitions. For example: ``A person walks, then runs, then jumps'',
   or ``A car drives through sand, then forest, then snow''.\\
4.~Make the temporal order explicit with connectors such as ``then''.\\
5.~Avoid abstract descriptions that a video generation model is
   unlikely to visualize. Focus on concrete, observable content.\\
6.~Ensure that the events are visually distinct from one another, and
   describe only the visually observable content that would appear in
   the video, avoiding narrative explanations, internal thoughts, or
   background stories.\\[0.5ex]
Now generate prompts that satisfy all the conditions above.
\end{quote}

\begin{figure*}[t]
  \centering
  \includegraphics[width=0.8\linewidth]{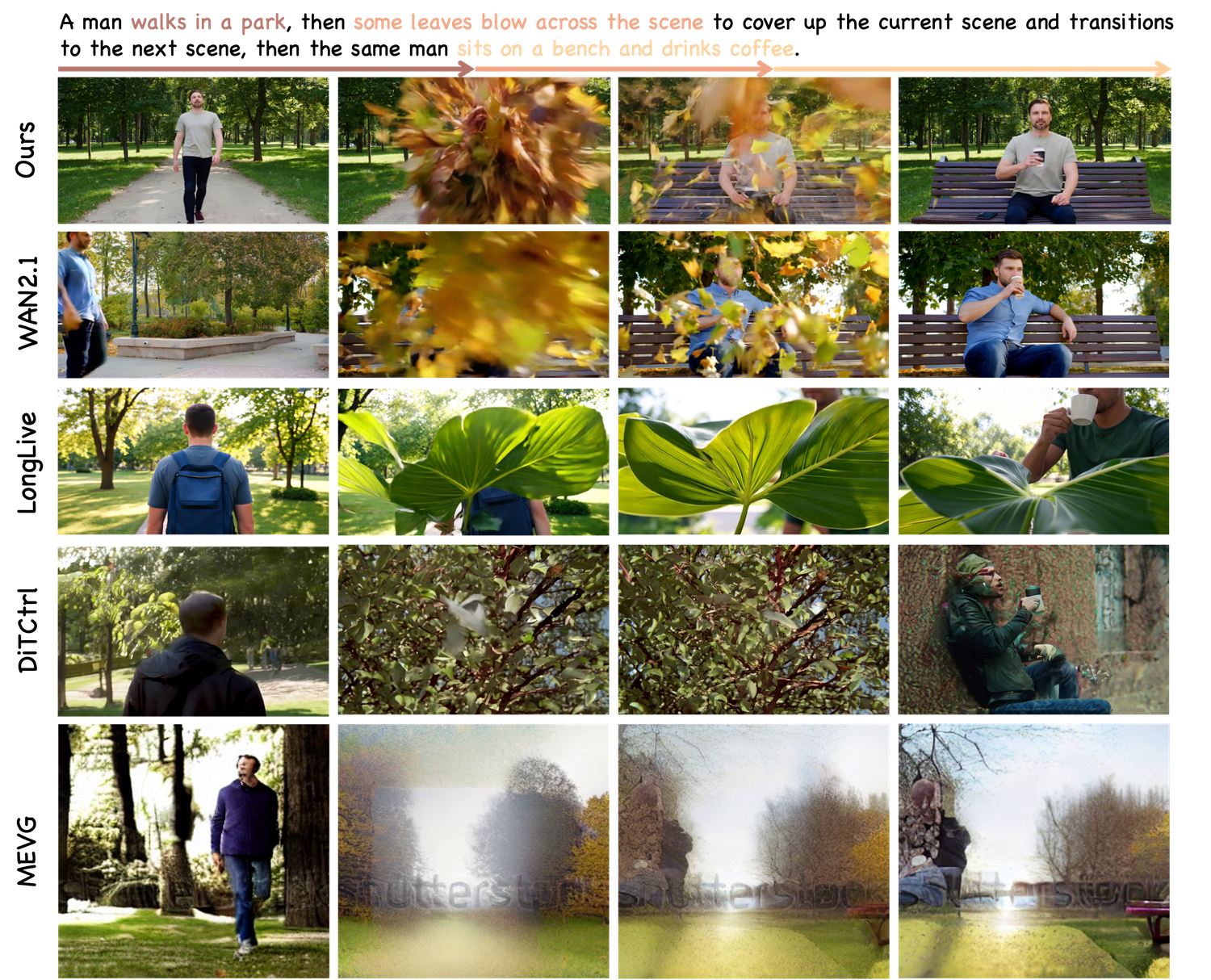}
  \caption{\textbf{Comparison of creative occluding transitions.} SwitchCraft produces in-shot occluding transitions that blend the
first and last scenes while preserving subject identity, whereas baselines
leave residual transition elements or degrade identity and scene
coherence.}
  \label{fig:trans_add1}
\end{figure*}

\begin{figure*}[t]
  \centering
  \includegraphics[width=0.8\linewidth]{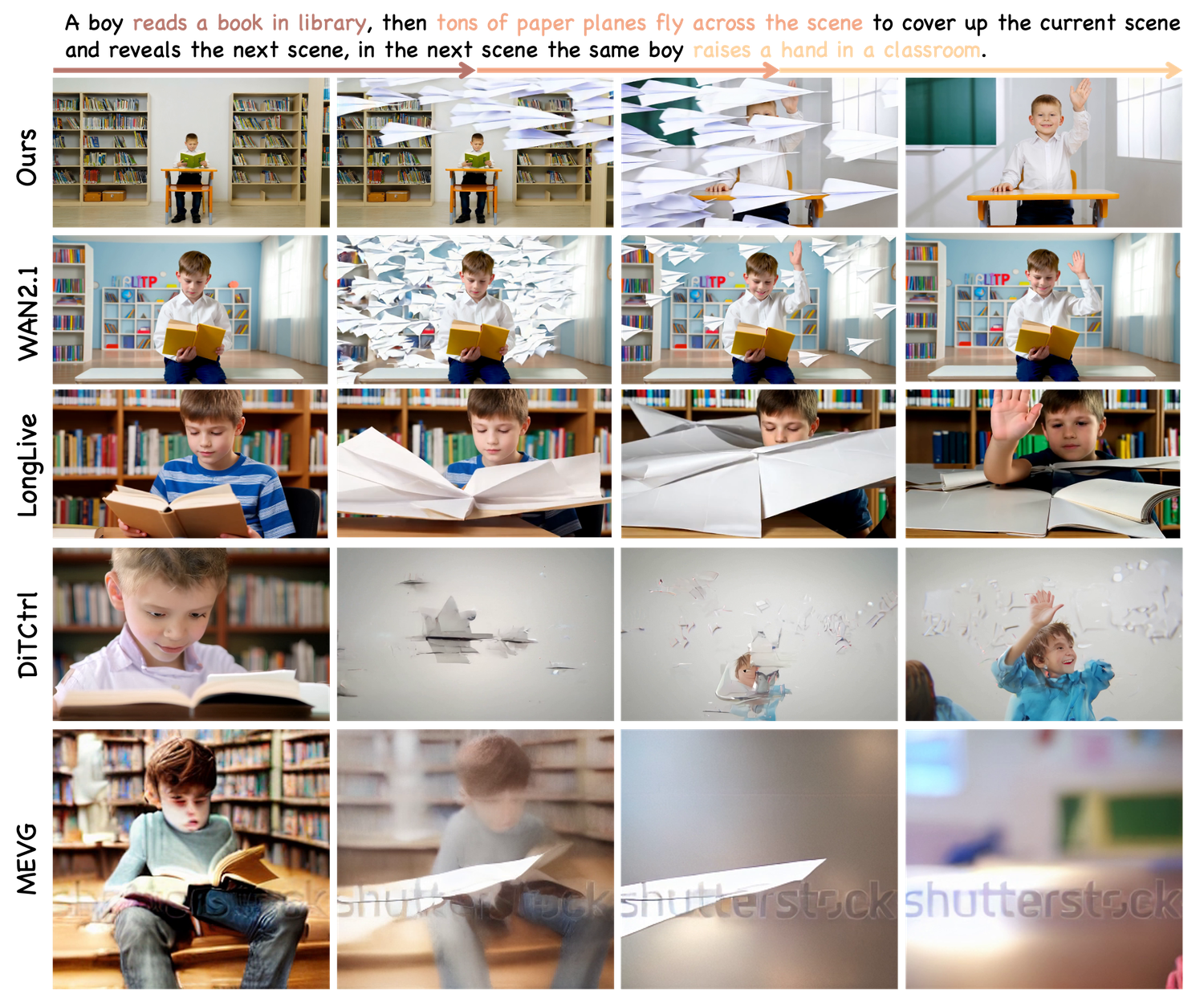}
  \caption{\textbf{Comparison of creative occluding transitions.} SwitchCraft produces in-shot occluding transitions that blend the
first and last scenes while preserving subject identity, whereas baselines
leave residual transition elements or degrade identity and scene
coherence.}
  \label{fig:trans_add2}
\end{figure*}

\begin{figure*}[t]
  \centering
  \includegraphics[width=\linewidth]{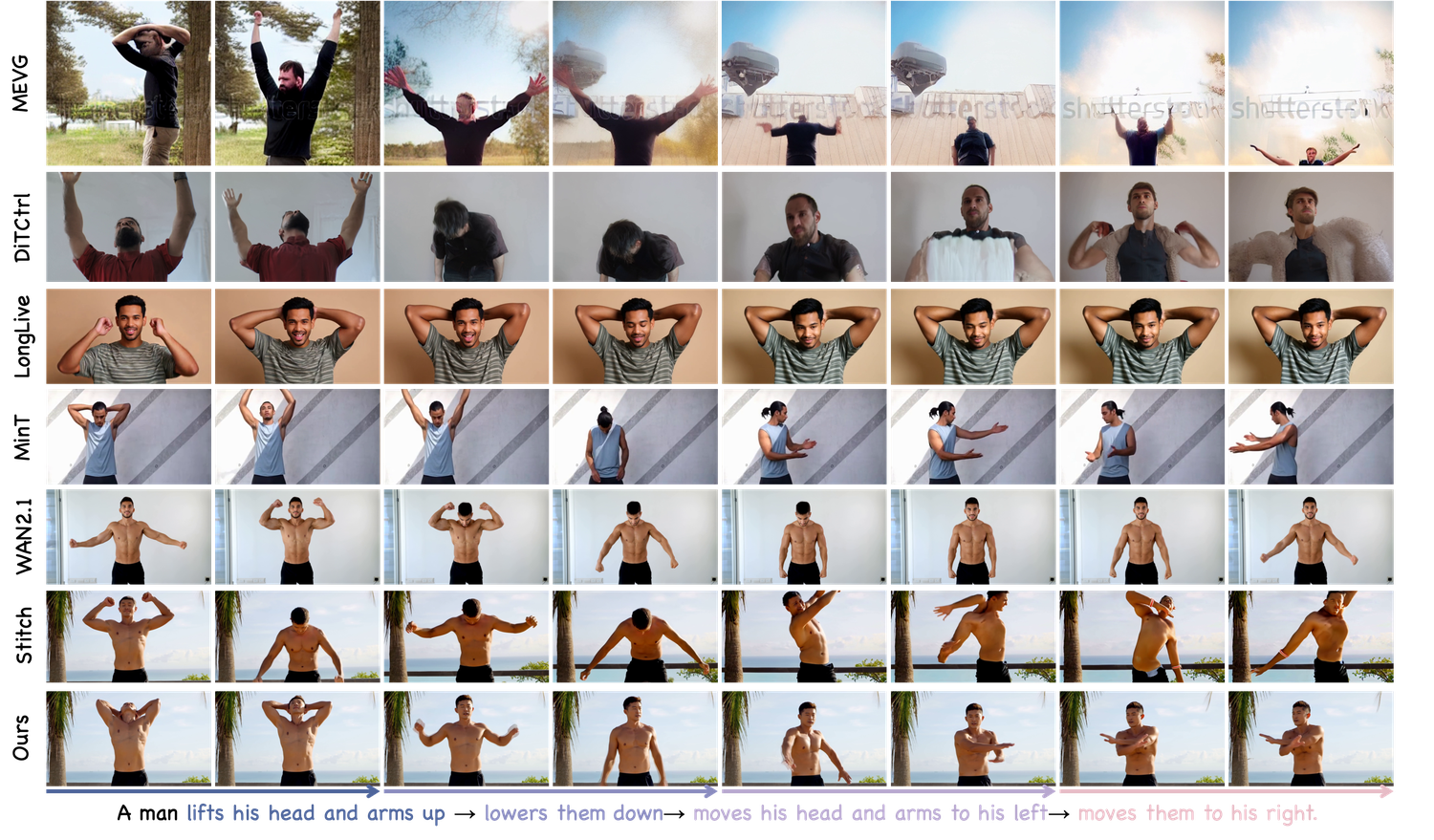}
  \caption{\textbf{Additional Qualitative comparison.} SwitchCraft executes all events in the intended order, prevents leakage and omission, and maintains subject and scene consistency. Baselines show omissions, drift across spans, or progressive quality decay.}
  \label{fig:comp_add}
\end{figure*}

\begin{figure*}[t]
  \centering
  \includegraphics[width=\linewidth]{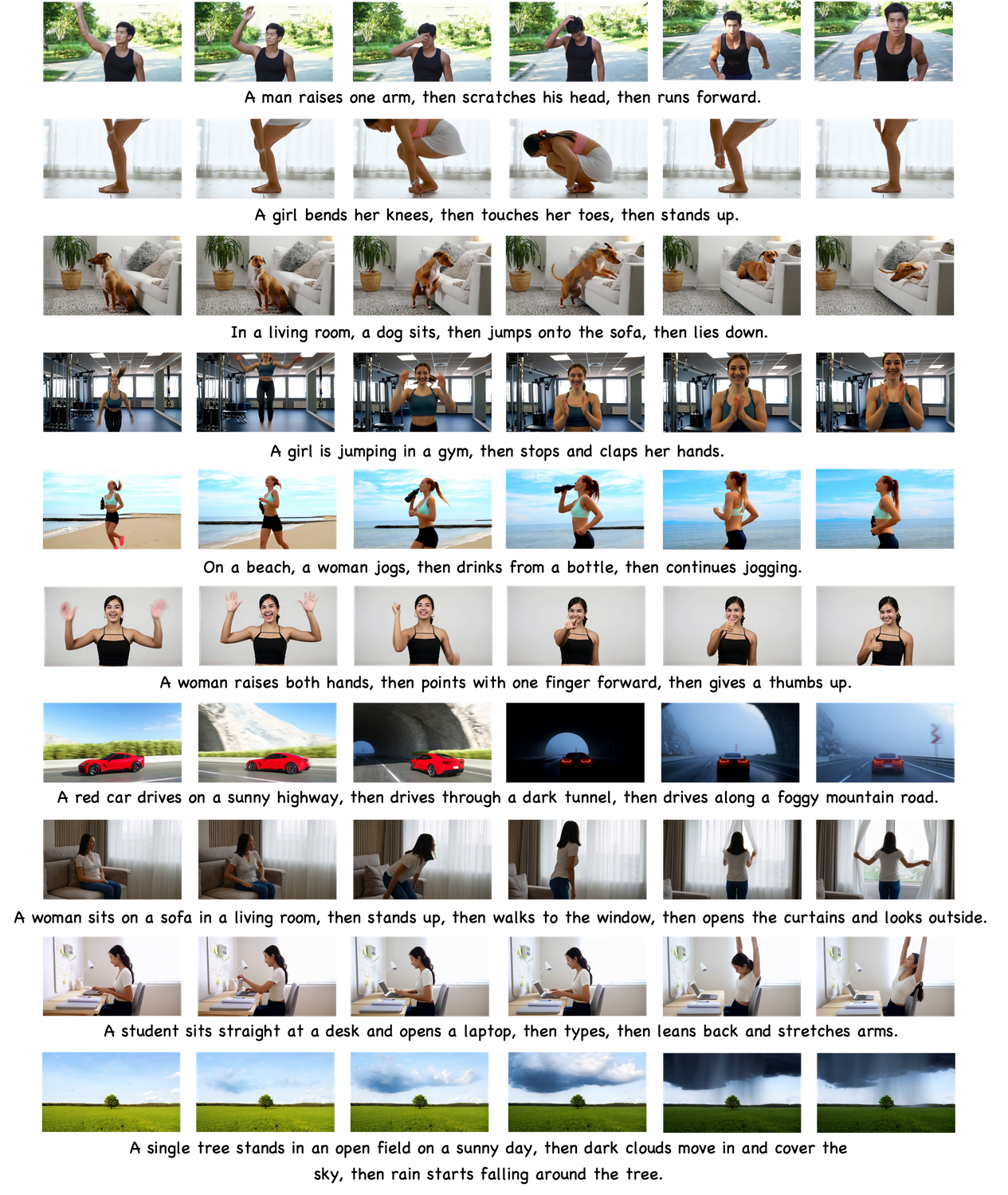}
  \caption{\textbf{Additional qualitative results.} For diverse multi-event prompts, SwitchCraft generates temporally coherent videos with smooth transitions between events, while preserving subject identity and global context.}
  \label{fig:add}
\end{figure*}

\section{Limitations}
\label{sec:limitations}

While SwitchCraft substantially improves controllability for multi-event text-to-video generation, it has several limitations as illustrated in Figure \ref{fig:limit}.

First, SwitchCraft inherits the strengths and weaknesses of the underlying video generation backbone. Our steering mechanism does not address generic limitations of current video diffusion models, such as fine-grained object fidelity under extreme motion, rare concepts, or perfect physical realism.  In such cases, our method can only guide the model within what the backbone is able to represent. For instance, if the base model cannot generate a specific complex action, our steering degrades gracefully to a generic approximation (e.g., failing to generate ``jumping jacks'' results in general jumping). 

Second, our method assumes that a multi-event prompt can be decomposed into a linear sequence of temporally ordered events and there is a single dominant subject that undergoes these events. While our framework can support multi-subject scenes by anchoring simultaneous actions to the same temporal span, our steering does not have an explicit mechanism to bind every event to a specific subject spatially. In such cases, correct subject-action attribution depends heavily on the base model's inherent spatial understanding, and the extracted event anchors may not perfectly match the user intent, causing actions to be softly mixed or swapped between subjects. 

We view these limitations as opportunities for further research on
more flexible multi-event video generation with subject-aware steering mechanisms, allowing for improved video quality and complex interactions in event sequences.

\section{Inference Time}
\label{sec:inference_time}

We report inference time for the Wan 2.1 T2V 14B backbone~\cite{wan2025} and
SwitchCraft on multi-event prompts with two, three and four events.
All videos are generated at a resolution of \(832 \times 480\) with
\(81\) frames. Timings are measured on a single NVIDIA A100 GPU.

For the Wan 2.1 baseline~\cite{wan2025}, inference time is essentially independent of the number of events described in the prompt, since there exists only one prompt and the diffusion sampler follows the same schedule regardless of the events mentioned in the content. SwitchCraft reuses the same sampler and backbone and adds a lightweight steering module that modulates cross attention according to event aligned queries. 

Table~\ref{tab:inference_time} summarizes the measured inference time for two, three and four event prompts for the Wan 2.1 baseline~\cite{wan2025} and SwitchCraft. Table~\ref{tab:latency_step} details the corresponding per-step inference latency in seconds, providing a breakdown of the computational overhead introduced by the EAQS and ABSS components compared to the base model across varying numbers of events. Note that EAQS+ABSS is applied only during the early diffusion steps and blocks.

\begin{table}[t]
  \centering
  \caption{Inference time per video (minutes) for Wan 2.1 T2V 14B~\cite{wan2025} and
  SwitchCraft on multi-event prompts of different lengths.}
  \label{tab:inference_time}
  \vspace{1.5pt}
  \begin{tabular}{@{}lcc@{}}
    \toprule
    \ Number of events
    & Wan 2.1~\cite{wan2025}
    & SwitchCraft \\
    \midrule
    2 & 15.2 & 17.6 \\
    3 & 15.2 & 19.7 \\
    4 & 15.2 & 22.3 \\
    \bottomrule
  \end{tabular}
\end{table}

\begin{table}[t]
\centering
\caption{Inference time per step (seconds) for the base model, EAQS, and EAQS+ABSS on multi-event prompts of different lengths.}
\label{tab:latency_step}
\vspace{1.5pt}
\small
\setlength{\tabcolsep}{4pt}
\renewcommand{\arraystretch}{1.05}
\begin{tabular}{lccc}
\toprule
 & \textbf{2 events} & \textbf{3 events} & \textbf{4 events} \\
\midrule
Base step (no edit)        &  & 18.66  \\
EAQS step (fixed $\alpha,\beta$) & 19.62 & 20.07 & 20.81 \\
EAQS+ABSS step & 22.32 & 26.83 & 30.19 \\
\bottomrule
\end{tabular}
\end{table}

\section{Analysis of Creative Occluding Transitions}
\label{app:occluding_transition}

In this section we illustrate why SwitchCraft can achieve creative
occluding transitions while other methods struggle. We define an occluding transition event as a single continuous video in which one event acts as a visual bridge between two scenes depicting the same subject. Typically, a subject performs an action in scene A, an occluder moves across the frame and temporarily occludes the view, and then the same subject continues a different action in scene B.
The transition must appear only within its temporal window, coherently
occlude scene A, and then disappear so that the final scene
contains only the new background and the persistent subject. 

As shown in Figure~\ref{fig:trans_add1} and Figure~\ref{fig:trans_add2}, SwitchCraft naturally handles this by operating within a single diffusion
trajectory and steering only event specific directions.
Global information is shared among all frames, and query steering gives the occluder a well defined temporal window in which it can
transition in and out smoothly, instead of creating hard cuts. While Wan 2.1 base model~\cite{wan2025} also preserves global information, it does not guarantee basic prompt alignment.

By contrast, LongLive~\cite{Yang2025LongLive} generates frames
autoregressively with cached memory.
Therefore, once a strong occluding object appears, it becomes a persistent feature in the recent memory, making it hard to fully preserve identity after heavy occlusion and the occluder lingers in later frames.

DiTCtrl~\cite{Cai2025DiTCtrl} achieves multi-event video as temporal
editing with smooth transitions by blending between separately generated
segments.
This design works well for transitions between similar scenes,
but when events differ substantially in content or layout, blending can
introduce ghosting transition elements and degrade identity consistency
in later segments.

MEVG~\cite{Oh2024MEVG} focuses on deliberate scene transitions by
blending or fading between separately generated event clips. However,
such clipwise blending cannot support a physical occluder that overlaps
both scenes in a single continuous shot and then disappears at the
correct time.

\section{Additional Qualitative Results}
\label{sec:add_qualitative}

In this section, we provide additional qualitative visualizations to
complement the main results, including a comparison across baselines in
Figure~\ref{fig:comp_add} and more examples generated by SwitchCraft in
Figure~\ref{fig:add}. 

Results show that SwitchCraft produces smoother
transitions between events and preserves high visual quality while staying faithful to the prompt. Additional SwitchCraft generated examples further illustrate that our steering framework can handle diverse multi-event prompts while maintaining clear event ordering and coherent global context.

\end{document}